# Adversarial Evaluation of Large Language Models for Patient-Friendly Radiology Report Summarization

## AUTHORS

Magdalini Paschali, PhD[1,2], Asad Aali, MS[1], Arogya Koirala, MS[1,2], Andrew Johnston, MD, MBA[1], Katherine McNamara, MD, PhD, MPhil[1], Richard Gaus, BS[1,3], Roxana Daneshjou, MD, PhD[4,5], David B. Larson, MD, MBA[1,2], Akshay S. Chaudhari, PhD[1,2,5]

## AFFILIATIONS

[1] Department of Radiology, Stanford University, Stanford, CA, USA.

[2] AI Development and Evaluation Lab (AIDE), School of Medicine, Stanford University, Stanford, CA, USA.

[3] Department of Computer Science, Technical University of Munich, Munich, Bavaria, Germany.

[4] Department of Dermatology, Stanford University, Stanford, CA, USA.

[5] Department of Biomedical Data Science, Stanford, CA, USA.

## CODE & RESOURCES

https://github.com/stanfordaide/adversarial-edits-llm-eval

## Abstract

**Purpose:**

To develop and validate a scalable adversarial evaluation framework for assessing the robustness of large language models (LLMs) that generate patient-friendly summaries of radiology reports.

**Materials and Methods:**

We introduced a clinically grounded adversarial framework that uses controlled term substitutions and impression removals across intensity levels. Using a radiologist-validated adversarial dictionary of 247 terms, we generated adversarial variants of 13,801 CT and MRI reports. For 300 reports (100 per dataset), six LLMs generated summaries from original and adversarial impressions using two prompt styles, yielding 18,000 evaluation pairs. Robustness was evaluated with an automated LLM-as-a-judge framework scoring factual faithfulness, sensitivity to adversarial changes, certainty fidelity, and hallucination rate. Failure mode analysis quantified vulnerabilities in incorporating clinical modifiers. Agreement with two radiologists was assessed on 105 cases. The code and radiologist-validated adversarial term dictionary used in this study are publicly available at https://github.com/stanfordaide/adversarial-edits-llm-eval.

**Results:**

GPT-5-mini achieved the highest factual faithfulness and certainty fidelity under both adversarial edits. Under term substitution, hallucinations occurred in 28.1-30.1% of GPT-5-mini summaries, compared with 35.3-36.4% for Claude Sonnet 4.5 and 41.4-41.9% for MedGemma. Impression-removal edits were handled more robustly than term substitutions. Even for the highest-scoring summarizer, substitutions of anatomical terms were not reflected in 23.7% of summaries, diagnostic certainty in 22.3%, and descriptor terms (e.g., “substantial”) in 21.2%. Judge-radiologist agreement (macro-F1: 0.55-0.56) was comparable to inter-radiologist agreement (0.53).

**Conclusions:**

Adversarial evaluation exposes clinically meaningful failure modes, including missed anatomical modifiers, altered diagnostic certainty, and omitted severity descriptors. This framework enables evaluation of how faithfully patient-facing LLMs preserve clinical modifiers and identifies semantic vulnerabilities relevant to quality control and future clinical deployment.

## Introduction

Recent legislation, such as the 21st Century Cures Act in the United States, has democratized patient access to their electronic health records, including the traditionally dense text of radiology reports [1]. While such transparency empowers patients, the specialized language of radiology often remains a barrier to comprehension, leading to potential anxiety and misinterpretation [2]. Large Language Models (LLMs) allow bridging this communication gap by translating complex medical terms into patient-friendly summaries [3-6]. Initial studies confirm that patients find these AI-generated summaries acceptable, highlighting their potential to enhance health literacy and patient-provider dialogue [7, 8].

Despite their linguistic fluency, the clinical deployment of LLMs for patient-facing applications introduces significant risks related to clinical accuracy and patient safety. These models can generate fluent outputs but may contain subtle yet clinically critical errors, including factual inaccuracies, misrepresentation of diagnostic uncertainty, and fabricated information, or "hallucinations" [9-11]. The absence of "ground truth" patient-friendly summaries makes traditional accuracy assessments difficult, while manual review by clinical experts, the current gold standard, is unscalable and unsustainable. Overcoming this validation bottleneck is a critical barrier to the safe, widespread clinical adoption of LLM-based summarization [12]. Current evaluation methods for clinical text summarization in radiology have relied on small-scale manual expert review for fact-checking and patient preference surveys for readability [13, 14]. While reliable, the need for physicians performing this evaluation renders it impractical at scale. Conversely, quantitative metrics originally developed for general text are scalable [15, 16], yet correlate poorly with clinical factuality and validity [12].

To overcome these limitations, we adopt adversarial testing, a common method in machine learning safety research, for evaluating the robustness of patient-facing text summarization [17-19]. Our adversarial testing approach (Figure 1) centers on creating controlled, clinically validated "adversarial edits"; systematic modifications in the original radiology reports (term substitutions, omissions, etc.) grounded in established adversarial principles. Using the introduced edits as known, controllable perturbations, we quantify the sensitivity of summarizer-LLMs to identify these adversarial edits. We apply our framework to over 13,800 CT and MRI reports across anatomical regions. We quantify performance of summarizer LLMs using an automated LLM-as-a-judge evaluation framework coupled with independent radiologist review. While prior work has used LLMs to generate adversarial modifications and fine-tuned LLMs for evaluation [20, 21], these approaches rely on LLMs to generate or grade the errors, which can themselves introduce inaccuracies. We therefore deliberately use non-LLM, radiologist-validated adversarial edits and off-the-shelf LLM judges to avoid this error propagation.

The purpose of this study is to develop and validate a scalable adversarial evaluation framework to assess the robustness of LLMs that generate patient-friendly radiology report summaries and use it to identify their characteristic failure modes. To support reproducibility, we publicly release our radiologist-validated adversarial dictionary, code, and prompts.

## Materials and Methods

### 1. Datasets

Our method used a diverse corpus of 13,801 radiology report impressions sourced from three distinct, publicly available datasets, namely Merlin [22], INSPECT [23], and MIMIC-IV [24] (Table 1). The datasets were de-identified and obtained under their respective data use agreements. For Merlin, we used the full official patient-level public test set (5,125 reports). For INSPECT, we used a patient-level test split (3,676 reports, ~16% of the full corpus) defined a priori; this split is released alongside our code. For MIMIC-IV, we restricted the radiology corpus to MRI reports, deduplicated to one report per patient, and selected the 5,000 reports with the longest impressions. On average, reports in Merlin contained 2.7 ± 1.4 impression points per report, INSPECT 2.8 ± 1.5, and MIMIC-IV 4.6 ± 2.2. This dataset selection ensured our findings were generalizable across CT and MRI, body regions, report lengths, and styles.

**Table 1: Overview of Datasets Used in the Study**

| Dataset | Modality | Body Region(s) | # of Reports |
|---|---|---|---|
| **Merlin [22]** | Abdominal CT | Abdomen | 5,125 |
| **INSPECT [23]** | Chest CTPA | Chest | 3,676 |
| **MIMIC-IV [24]** | MRI | Spine, Brain/Head, Breast, Knee, etc. | 5,000 |
| **Total** | | | 13,801 |

### 2. Adversarial Edit Generation

To evaluate the robustness of LLM-generated summaries, we designed an automated framework to modify original radiology impressions using two adversarial edit strategies: Term Substitution and Impression Removal. For each strategy, we generated edits at different levels of modifications to probe the limits of model performance.

**Term Substitution**

This strategy involved substituting clinically relevant terms that alter the original radiology report's meaning. Using Natural Language Processing (NLP), we extracted medical terms and grouped them into eight categories: pathology, descriptors, anatomy, location, confidence terms, study types, measurements, and findings (Figure 2) [25]. These categories were drafted by the study team, iteratively refined with a radiologist, and reviewed by a board-certified radiologist. Each term was matched with alternatives that, when used as a replacement in a radiology report, would change its meaning. For example, the descriptor term "mild" could be replaced with alternatives such as "severe" or "marked". Our radiologist-validated dictionary contained 247 terms, including multi-word phrases. The adversarial dictionary deliberately spans a spectrum of clinical plausibility. Categories such as confidence terms, location, and descriptors (e.g., substantial → small) yield clinically plausible adversarial impressions that mirror real-world variability and probe whether LLM summarizers faithfully propagate fine-grained clinical details, including the adversarial edits, into the patient-friendly summaries. Substitutions in categories such as pathology and anatomy ("embolus" → "atelectasis"; "adenoma" → "adenocarcinoma") intentionally probe model behavior at the edges of the input distribution, including errors arising from speech-recognition or dictation and error modes of LLMs that could hallucinate terms [26]. This spectrum allows the framework to characterize summarizer robustness across both realistic and worst-case perturbations. For the measurements category, each eligible numerical measurement was altered with a probability of 0.5 (e.g., 1.5 cm → 3.3 mm). The detailed methodology for the adversarial term dictionary's construction is available in the Supplementary Materials; the complete dictionary is released as Supplementary Data.

For each report, the substitution process was performed with three levels of edit intensities:

- Level 1: Exactly one term from a single semantic category was substituted.
- Level 2: Two terms from each of two semantic categories were substituted.
- Level 3: Four terms from each of four semantic categories were substituted.

These three levels increase the perturbation intensity approximately geometrically (up to 1, 4, and 16 total substitutions, respectively) to span minimal to substantial edits.

If a report contained fewer eligible terms than the target number of substitutions for a given level, the algorithm attempted substitutions using alternative categories or multi-word phrases to approach that target. If the available terms and phrases still cannot meet the target, the report is retained with the maximum number of substitutions that could be made. To ensure comprehensive testing across the full breadth of the dictionary and avoid oversampling common terms, the algorithm maintained a global usage count for each term. When selecting a term for substitution, it prioritized those that had been used least frequently in previous alterations.

**Impression Removal**

This strategy evaluated summary completeness by removing content from the impression section of original reports. When a report had more than one impression point (the numbered statements), we removed one or more entire impression points. When a report had a single impression point, we removed a sentence or a keyword from within that point instead. The impression text was first segmented into constituent numbered points using a regular expression (e.g., "1. ", "2. "). Each point was then scored as the unweighted count of medical keywords it contained, drawn from a per-dataset vocabulary built with a hybrid NLP pipeline that combined scispaCy [27], biomedical named entity recognition, and rule-based patterns for radiological nomenclature (e.g., measurements, staging). The scale, therefore, ranged from 0 to N keywords per point, with "highest" denoting the most keyword-dense point.

Removal occurred at two levels of intensity:

- Level 1: The single impression with the highest medical keyword score was removed.
- Level 2: Up to three of the highest-scoring impressions were removed.

When the impression was a single point with several sentences, we applied the same medical-keyword scoring to each sentence and removed the highest-scoring sentence(s), breaking ties by position. When the impression was a single sentence, we removed one of its medical keywords, chosen at random.

## 3. Patient-Friendly Summary Generation

Following the creation of adversarial edits, patient-friendly summaries were generated for original and adversarial impressions using six LLMs (GPT-5-mini, Claude Sonnet 4.5, MedGemma, Llama 3.1, Llama 3.3, and Qwen 3; Table 2) to analyze the impact of model size, defined as the number of trainable parameters, and domain-specific training on robustness.

For each model, we generated summaries using two distinct prompt formulations to evaluate the influence of prompt engineering (full text available in Supplementary Materials). A "detailed prompt" assigned the LLM the persona of a medical communication specialist with explicit instructions regarding tone, content, and evaluation criteria. A "concise prompt" provided a direct, single-sentence instruction to create a patient-friendly summary. Summaries and judgments were generated with a sampling temperature of 1.0 for the proprietary API models (GPT-5, GPT-5-mini, and Claude Sonnet 4.5) and 0.8 for the open-weight models (Llama 3.1, Llama 3.3, Qwen 3, MedGemma). Each query was executed once.

**Sampling and Experimental Combinations.** To be able to apply the full set of experimental permutations across summarizers, prompts, adversarial edits, and their associated levels, we randomly sampled 100 reports from each dataset. Each sampled report impression section was summarized by six LLMs using two prompting strategies and evaluated under Term Substitution and Impression Removal on different levels (three levels for Term Substitution and two levels for Impression Removal). Across all combinations (3 datasets × 100 reports × 6 summarizers × 2 prompts × [3 term-substitution + 2 impression-removal levels]), we examined a total of 18,000 evaluation pairs (36,000 total summaries).

**Table 2: Language Models Used for Summary Generation and as Judge**

| Model Name | Summarizer | Judge | Parameters | Type | Domain |
|---|---|---|---|---|---|
| **GPT 5 [34]** | | ✓ | N/A | Proprietary | General |
| **GPT 5-mini [34]** | ✓ | | N/A | Proprietary | General |
| **Claude Sonnet 4.5 [35]** | ✓ | | N/A | Proprietary | General |
| **Llama 3.1 [36]** | ✓ | | 8B | Open-Source | General |
| **MedGemma [37]** | ✓ | ✓ | 27B | Open-Source | Medical |
| **QWEN 3 [38]** | ✓ | | 4B | Open-Source | General |
| **Llama 3.3 [36]** | ✓ | | 70B | Open-Source | General |

## 4. Automated Evaluation Framework (LLM-as-a-Judge)

To assess summary quality, we implemented an automated framework using GPT-5 as the primary LLM judge and MedGemma as an alternative. GPT-5 was selected as the primary judge because, at the time of evaluation, it was the most recent state-of-the-art proprietary OpenAI model with strong benchmark performance and was not used as a summarizer in this study, avoiding self-evaluation bias. MedGemma was included as a secondary medically adapted, open-source judge. Our findings for GPT-5 judge are reported in the Results Section, while the results for MedGemma judge are available in the Supplementary Material (Table S2 and Figures S1-S3), along with the prompts used by the judges. The evaluation for each impression-summary pair was conducted in a two-stage process:

1. Structured Fact Extraction: The judge model was first prompted to extract clinical findings into a structured schema, identifying condition, location, severity, and certainty of each finding from both original and adversarial impressions. This intermediate structured representation decomposed free-text reports into comparable clinical elements, enabling more reliable detection of fine-grained inconsistencies and providing a more stable basis for evaluation [28].
2. Comparative Judgment and Scoring: The judge compared the structured facts and summaries to evaluate the adversarial summary across four dimensions:

   - **Factual Faithfulness**: Preservation of unaltered original facts.
   - **Sensitivity to Changes**: Reflection of specific adversarial edits.
   - **Certainty Fidelity**: Maintenance of diagnostic certainty levels.
   - **Hallucination Rate**: Presence of findings absent in source documents.

For simplicity, Factual Faithfulness, Certainty Fidelity, and Hallucination Rate were treated as binary tasks, reflecting whether facts or diagnostic certainty were preserved and whether hallucinations were present. Sensitivity to changes was graded on a 3-point scale to capture partially reflected edits. Factual faithfulness required preservation of four attributes associated with a finding (condition, location, severity, certainty). A summary missing or altering any of the attributes was scored as unfaithful. Clinically equivalent paraphrases, including those that conveyed quantitative measurements through qualitative descriptors (e.g., "mild enlargement" rather than a specific size), were considered faithful provided all four attributes were preserved. Both stages were executed as separate, sequential API calls to the judge LLM, with the structured findings produced in Stage 1 passed as input to Stage 2. Finally, we performed a failure mode analysis by comparing judge-detected alterations against ground-truth edits to calculate miss rates across categories and individual terms. Specifically, the judge was prompted to output the list of detected alterations between the original and adversarial summaries (the "Alterations" field in the Stage 2 prompt; Supplementary Materials). For each evaluation pair, these detected alterations were compared against the ground-truth substitutions using regular-expression matching with allowance for morphological variants (plurals, common inflections). A "miss" was recorded when a ground-truth substitution did not appear among the detected alterations. Miss counts were aggregated separately by term category (e.g., anatomy, descriptors) and by individual term to produce the rates reported in Figure 5.

Two resident radiologists (Rad A: 4th year, Rad B: 3rd year) independently rated 105 impression-summary pairs, sampled evenly across datasets using the same rubric to validate the automated LLM-as-a-judge framework. Rad A also participated in validation of the fixed adversarial dictionary, whereas Rad B had no role in dictionary development. All reader-study summaries were generated by GPT-5-mini using the concise prompt; readers were blinded to model identity. Readers had access to the original and adversarial summaries, as well as the list of altered terms, as required to assess sensitivity to changes. The reader-study sample size (n=105) was selected a priori with a precision-based calculation to balance expert-review feasibility with statistical precision. Agreement between radiologists and the GPT-5 judge was quantified using symmetric macro F1-scores, as reported in the Results section. Regarding the F1-score, n=105 yields an approximate worst-case 95% confidence interval half-width of approximately ±10 percentage points for proportions, providing sufficient precision to assess agreement while maintaining a tractable radiologist workload. Cohen's κ was not selected as the primary agreement measure because the evaluation labels exhibit substantial class imbalance, a setting in which κ is known to be unstable due to the prevalence paradox [29]. To characterize the observed hallucinations, we manually labeled the individual hallucinations in a reader-study subset of reports that were flagged as containing hallucinations by at least two of the three assessors (the two radiologists and the GPT-5 judge). Within the flagged reports, each hallucination was assigned, using categories derived retrospectively from the observed hallucinations, to one of eight types: unsupported follow-up recommendations, added confidence qualifiers, added reasoning, added details, fabricated descriptors, fabricated location, fabricated pathology, and fabricated anatomy; a report could contain multiple types. These hallucination categories are not intended to be exhaustive; other report types or settings may exhibit additional types.

**Statistical Analysis.** Differences between detailed and concise prompts were assessed using two-sided nonparametric permutation tests within each summarizer and edit condition (20,000 permutations); p-values are reported directly in the Figures. Uncertainty in F1-scores was estimated via bootstrap resampling of cases (800 bootstrap samples), and results are reported as mean ± standard deviation. Repeated-measures models corroborating these results are reported in the Supplement (Supplementary Tables S4 to S6). The code and radiologist-validated adversarial term dictionary are publicly available at https://github.com/stanfordaide/adversarial-edits-llm-eval.

## Results

### 1. Factual Faithfulness and Certainty Fidelity

Throughout the Results, paired values separated by a slash denote the detailed and concise prompts, respectively. Under Term Substitution, GPT-5-mini achieved a mean factual faithfulness score of 0.851 (95% CI: 0.826-0.873) with detailed prompts and 0.859 (95% CI: 0.835-0.880) with concise prompts. Claude Sonnet 4.5 followed with 0.756 and 0.703, while MedGemma achieved 0.670 and 0.713. Scores for Llama 3.1 and Qwen 3 were below 0.50. Under Impression Removal, scores were higher across all models: GPT-5-mini reached 0.942/0.950, Claude Sonnet 4.5 scored 0.913/0.883, and MedGemma reached 0.832/0.870 for detailed/concise prompts (Figure 3b). Nonparametric permutation tests showed that prompt variation did not affect factual faithfulness for most summarizers, except for Llama 3.3 ($p<0.01$) and Claude Sonnet 4.5 ($p<0.05$) under term substitution, and Qwen 3 ($p<0.05$) under impression removal (Figure 3b).

Across both edit types, certainty fidelity closely tracked factual faithfulness. For GPT-5-mini, certainty fidelity remained numerically close to factual faithfulness under both Term Substitution (0.849/0.848 for detailed/concise prompt) and Impression Removal (0.893/0.897 for detailed/concise prompt) (Figure 3a).

### 2. Sensitivity to Adversarial Edits

For Term Substitution, summaries from GPT-5-mini fully reflected edits in 39.9% (359/900) / 39.0% (351/900) of cases and partially reflected them in 42.0% (378/900) / 43.4% (391/900) for detailed/concise prompts. Under Impression Removal, sensitivity was higher: GPT-5-mini reflected changes in 89.8% (539/600) / 86.8% (521/600) of cases (detailed/concise), while Claude

Sonnet 4.5 fully reflected the edits in 90.2% (541/600) / 88.7% (532/600) of summaries (detailed/concise). Differences between prompt styles for sensitivity were observed only for Llama 3.1 ($p<0.05$), Qwen 3 ($p<0.001$), and MedGemma ($p<0.05$) under term substitution (Figure 4a).

### 3. Hallucination Rates

Under Term Substitution, GPT-5-mini exhibited hallucination rates of 28.1% (253/900) / 30.1% (271/900) for detailed/concise prompts. Claude Sonnet 4.5 showed rates of 35.3% (318/900) / 36.4% (328/900), and MedGemma showed 41.9% (377/900) / 41.4% (373/900) for detailed/concise prompts. Rates exceeded 60% for Qwen 3 (71.2% [641/900] and 61.1% [550/900]) and Llama 3.1 (60.7% [546/900] and 66.2% [596/900]) for detailed/concise prompts. For Impression Removal, rates were lower: GPT-5-mini hallucinated in 14.0% (84/600) / 18.2% (109/600) of cases, while Claude Sonnet 4.5 and MedGemma ranged between 23.3% (140/600) and 27.5% (165/600) (Figure 4b). Prompt-dependent differences in hallucination rate were noted for Qwen 3 ($p<0.001$ under term substitution and $p<0.01$ under impression removal) and Llama 3.1 ($p<0.05$ under term substitution; Figure 4b).

### 4. Failure Mode Analysis

Analysis of category-level failures under Term Substitution identified systematic vulnerabilities in reflecting clinical modifiers. Failure rates were highest for terms in the categories of Anatomy (23.7% [248/1,048] for GPT-5-mini; 26.6% [279/1,048] for Claude Sonnet 4.5), Descriptors (21.2% [331/1,558]; 23.6% [367/1,558]), and Confidence Terms (22.3% [85/382]; 19.6% [75/382]). In contrast, pathology terms were missed least frequently (8.1% [102/1,256] for GPT-5-mini; 11.7% [147/1,256] for Claude Sonnet 4.5) (Figure 5a).

Term-level failure analysis highlighted specific high-risk words, including descriptors such as “suspicious” (failure rate: 52.2% [24/46] for GPT-5-mini; 58.7% [27/46] for Claude Sonnet 4.5) and “substantial” (23.7% [54/228]; 26.8% [61/228]), as well as location terms such as “adjacent” (26.9% [21/78]; 32.1% [25/78]) and “low” (43.2% [19/44] for GPT-5-mini) (Figure 5b).

### 5. Radiologist Reader Study - LLM-as-a-Judge Validation

Table 3 summarizes the macro F1-scores against radiologists and radiologist inter-rater agreement. For factual faithfulness, GPT-5 achieved macro F1-scores of 0.52 against Radiologist A and 0.59 against Radiologist B (inter-radiologist agreement: 0.54). Agreement for certainty fidelity was 0.52-0.54, compared to 0.50 between radiologists. The highest agreement occurred in sensitivity to changes (0.59-0.61). For hallucination rate, F1-scores were 0.52 for GPT-5 against both radiologists (inter-radiologist agreement: 0.46). The overall macro F1-score was 0.55-0.56 for GPT-5 vs. radiologists and 0.53 for inter-radiologist agreement. Of the 105 reader-study reports, 66 were flagged as containing at least one hallucination by at least two of the three independent assessors (the two radiologists and the GPT-5 judge). These 66 reports contained 100 individual hallucinations in total (mean 1.5 ± 0.5 per report, range 1-3), with many reports exhibiting more than one hallucination type. Manual categorization of these 100 hallucinations identified unsupported follow-up recommendations (n=45), added reasoning (n=15), added confidence qualifiers (n=12), and added details (n=11) as the most frequent types, followed by fabrication of descriptors (n=9), location (n=6), pathology (n=2), and anatomy (n=0).

**Table 3: Inter-Rater Agreement Between GPT-5 and Radiologists for Adversarial Patient-Friendly Summary Evaluation (n=105).** F1 scores represent symmetric macro F1 (mean ± SD from 800 bootstrap iterations).

| Metric | Rater Pair | F1-Score |
|---|---|---|
| **Factual Faithfulness** | GPT-5 *vs.* Rad A | 0.52 ± 0.05 |
| | GPT-5 *vs.* Rad B | 0.59 ± 0.05 |
| | Rad A *vs.* Rad B | 0.54 ± 0.06 |
| **Certainty Fidelity** | GPT-5 *vs.* Rad A | 0.54 ± 0.06 |
| | GPT-5 *vs.* Rad B | 0.52 ± 0.05 |
| | Rad A *vs.* Rad B | 0.50 ± 0.05 |
| **Sensitivity to Changes** | GPT-5 *vs.* Rad A | 0.61 ± 0.06 |
| | GPT-5 *vs.* Rad B | 0.59 ± 0.07 |
| | Rad A *vs.* Rad B | 0.61 ± 0.07 |
| **Hallucination Rate** | GPT-5 *vs.* Rad A | 0.52 ± 0.05 |
| | GPT-5 *vs.* Rad B | 0.52 ± 0.05 |
| | Rad A *vs.* Rad B | 0.46 ± 0.05 |
| **Overall** | GPT-5 *vs.* Rad A | 0.55 ± 0.04 |
| | GPT-5 *vs.* Rad B | 0.56 ± 0.04 |
| | Rad A *vs.* Rad B | 0.53 ± 0.06 |

## Discussion

In this study, we systematically evaluated the robustness of large language models (LLMs) in generating patient-friendly summaries of radiology impressions under adversarial conditions. By introducing controlled term substitutions and impression removals, and by comparing six summarizers across three datasets and two prompting styles, we assessed performance on factual faithfulness, certainty fidelity, sensitivity to edits, hallucination rates, and miss patterns. Our findings carry several implications for the integration of generative AI in radiology-reporting workflows.

First, the high factual faithfulness and low hallucination rates observed for proprietary frontier models such as GPT-5-mini and Claude Sonnet 4.5 underscore that modern LLMs can faithfully rephrase radiology impressions for patient consumption. Prompt style had no significant main effect on any of the four metrics in the odds-ratio analysis, although the detailed prompt led to a small reduction in the mean sensitivity score. The interaction between the prompt style and the summarizer was significant across all four metrics. Thus, the effect of detailed versus concise prompting depended on the model rather than on the prompt alone (Supplementary Tables S4 and S5). The two prompt formulations differed in the amount of guidance they gave rather than in length as an independent factor: the detailed prompt assigned a persona and explicit instructions, whereas the concise prompt was a single-sentence request. However, the residual variability, particularly among the open-source models evaluated, highlights that model size, training-data domain, and fine-tuning were potential contributors to reliability among the six models we tested. For the models evaluated here, our results suggest that model selection remains a critical step and that not all the tested LLMs were equivalent in a clinical context.

Summaries that achieved high factual faithfulness did not uniformly achieve high scores across the other evaluation dimensions, underscoring the need for multi-dimensional assessment. In the Supplementary, we present representative cases in which summaries can achieve high scores on some metrics while failing on others. Factual faithfulness and certainty fidelity are related but not interchangeable. Our radiologist-guided evaluation criteria incorporated the nuances and multi-dimensional nature of evaluating patient-friendly summaries. For example, a summary may preserve factual content while subtly altering the level of certainty, which can meaningfully change patient interpretation and downstream clinical decision-making. This distinction highlights the importance of evaluating calibration of certainty explicitly, rather than inferring it from factual correctness alone. Furthermore, hallucinations in this study were defined as the introduction of any new clinical finding, description, or recommendation absent in the impressions. This definition focused on content that could mislead patients by introducing unsupported medical claims. The hallucination rates showed that even when summaries have high factual faithfulness and certainty fidelity scores, they may introduce information not originally included in the source report. Manual categorization of hallucinations indicated that most were additions of follow-up recommendations, confidence qualifiers, or supportive reasoning rather than fabrications of clinical entities such as pathology or anatomy, suggesting that quality-control measures targeting added recommendations and unwarranted explanations would capture most patient-relevant hallucinations.

Our failure mode analysis provided additional insight into systematic failure modes of LLM summarizers. Higher miss rates for descriptors, anatomical modifiers, and confidence terms indicate that summarizers disproportionately struggled with clinically nuanced language rather than pathological entities. From a radiological perspective, these categories encode information that modulates urgency, localization, and diagnostic interpretation. Their omission in patient-facing summaries may therefore downplay clinically relevant nuance even when primary findings are preserved [30]. These results suggest that robustness improvements should prioritize fine-grained semantic elements, instead of focusing on disease entities. From a clinical governance perspective, such findings could inform quality-control checklists for generated summaries (e.g., a “severity/region modifier present” rule) to trigger human review based on specific trigger types. The asymmetry between term-substitution and impression-removal performance reflects the granularity of the perturbations: impression removal is a coarse perturbation that leaves a salient absence in the source, whereas term substitution requires the summarizer to preserve fine-grained modifiers and confidence terms during paraphrasing.

Coupling our automated analysis with two radiology experts, we benchmarked the GPT-5 judge against inter-radiologist variability. Hallucination detection showed the lowest agreement overall, including between radiologists. Although we defined “hallucination” in the prompt and rating instructions, determining what constitutes a hallucination is inherently subjective; as a result, GPT-5 and the radiologists may differ in their interpretations. The moderate inter-radiologist agreement (F1 0.53) and judge-radiologist agreement (0.55-0.56) reflect the inherent subjectivity of evaluating patient-friendly radiology summaries. This task requires balancing clinical accuracy with patient-facing readability, and comparable variability is well documented in open-ended text-generation evaluation [12]. Future work will focus on improved rater calibration and consensus-building protocols to reduce variability. Moreover, cross-judge analysis with MedGemma, an open-source medically-adapted alternative judge, yielded consistent model rankings and failure-mode patterns (Supplementary Materials), with a marginal self-preference that did not affect the qualitative conclusions; primary results are reported with GPT-5 to eliminate any self-evaluation effect.

Despite these strengths, some limitations of our work merit consideration. Agreement was lower for factual faithfulness, certainty fidelity, and hallucination detection, even between the two radiologists. Our adversarial edits, though clinically motivated, are synthetic and may not fully capture the breadth of real-world reporting variability, such as multi-finding errors, free-text reformulations, or omissions unrelated to our adversarial perturbations. Additionally, our sampling framework, consisting of 100 impressions per dataset, was optimized for combinatorial design but may not reflect rare findings. Our focus on the impression section reflects its role as the distilled, clinically actionable summary of a radiology report and follows prior patient-facing summarization systems (e.g., RadGPT [31]). Nonetheless, extending adversarial perturbations to the findings section of radiology reports would provide a more complete assessment and is an important direction for future work. Looking ahead, future research should evaluate real-world integration: how summarizers affect patient comprehension and follow-up adherence; how adversarial perturbations translate into actual patient-communication risks; and how a human-LLM co-workflow can best be managed in production environments. The framework could also be extended to other clinically relevant error sources, such as speech-recognition errors from dictation systems [32, 33].

Guided by the specific failure modes identified in our analysis (e.g., frequently missed descriptors, anatomical modifiers, and confidence terms), future work should also explore advanced prompt engineering techniques, such as chain-of-thought prompting and few-shot exemplars, to improve standardization and consistency of both summarizer and judge outputs. Additionally, this study focused on clinical accuracy and did not directly evaluate the readability or patient-friendliness of the generated summaries; future work should incorporate patient comprehension assessments to complement adversarial robustness evaluation.

In conclusion, our adversarial evaluation highlights both the promise and the limitations of modern LLMs for patient-friendly radiology summary generation. While high-performing models can faithfully translate impressions, semantic nuance remains a vulnerability. Our framework measures how faithfully models preserve fine-grained clinical modifiers at scale and provides a foundation for the prospective, patient-facing studies that will guide safe clinical deployment.

# Figures

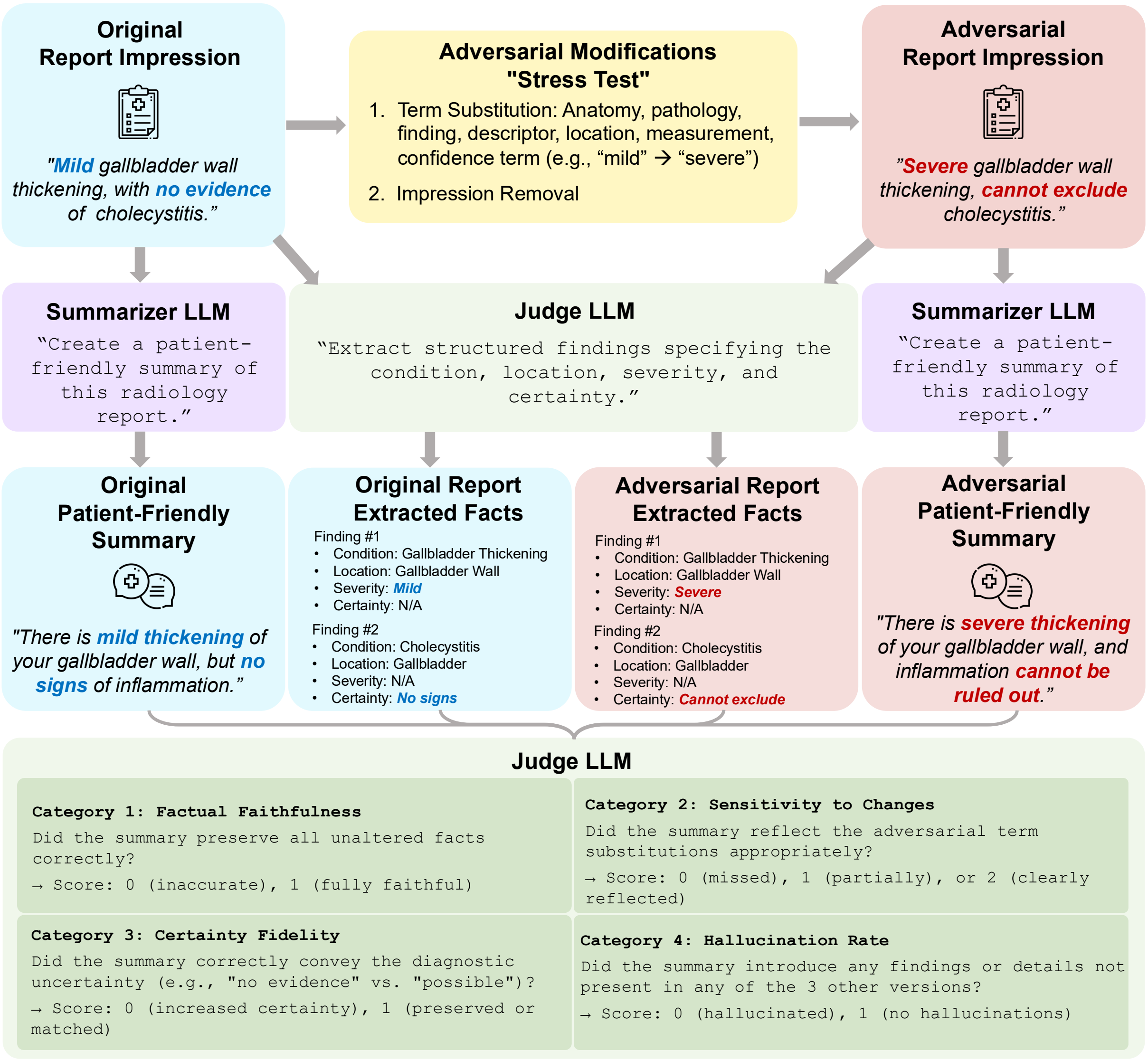

**Figure 1.** Overview of the adversarial evaluation framework for patient-friendly radiology summaries. Radiology report impressions undergo adversarial modification using predefined term substitutions or impression removal to create adversarial impressions. Summarizer LLMs generate patient-friendly summaries for both original and adversarial impressions, while the judge LLM extracts structured findings from the original and adversarial impressions. Using the extracted facts and summaries, it evaluates the adversarial patient-friendly summary. The judge scores factual faithfulness, certainty fidelity, sensitivity to changes, and hallucination rate, enabling automated detection of clinically relevant summarization errors. For the case shown, the original impression states "Mild gallbladder wall thickening, with no evidence of cholecystitis," and the adversarial impression substitutes "mild" → "severe" and "no evidence" → "cannot exclude." The adversarial patient-friendly summary states, "There is severe thickening of your gallbladder wall, and inflammation cannot be ruled out." The Judge LLM scores this summary as: Factual Faithfulness = 1 (the unaltered fact "gallbladder wall thickening" is preserved); Sensitivity to Changes = 2 (both substitutions are clearly reflected); Certainty Fidelity = 1 (the shift from "no evidence" to "cannot exclude" is faithfully conveyed); and Hallucination Rate = 0 (no findings absent from all source documents are introduced).

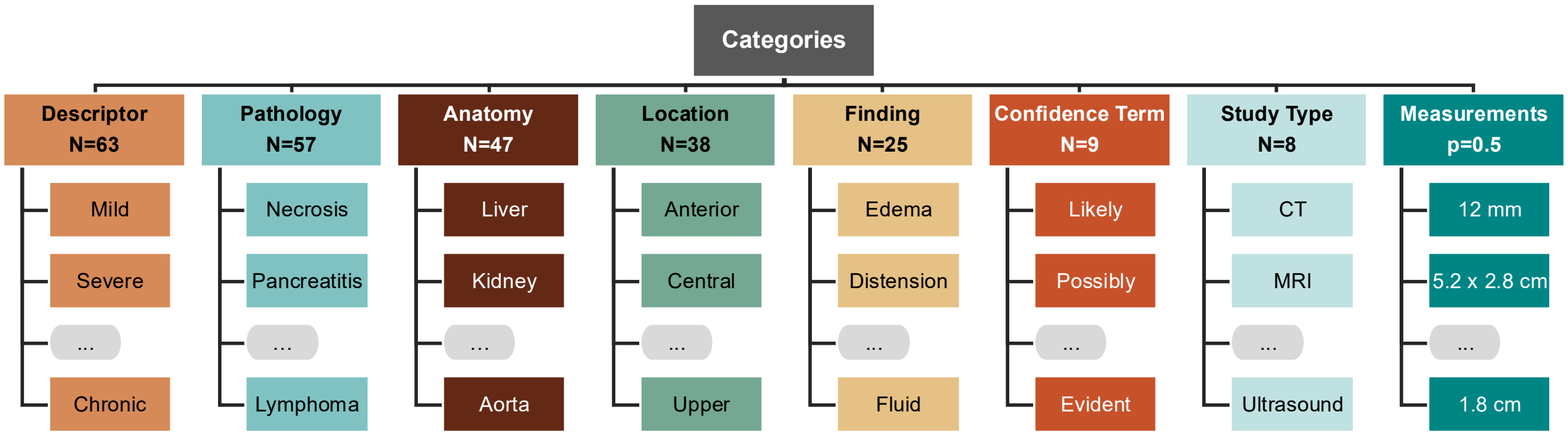

**Figure 2.** Taxonomy of adversarial edit categories. In collaboration with a radiologist, we define eight categories of targeted adversarial edits applied to radiology report impressions, including descriptors, pathology terms, anatomy, location, findings, confidence terms, study type, and measurements. N indicates the number of terms extracted for each category. For the Measurements category, numeric values and units (mm or cm), including multi-dimensional expressions, are randomly altered with probability $p$=0.5. All other category terms are substituted using our term alternatives, which are semantically related yet non-synonymous, yielding clinically distinct edits. For example, "mild" and "severe" are semantically related (both are severity descriptors) but not synonymous (they convey opposite meanings), making "severe" a valid adversarial substitute for "mild."

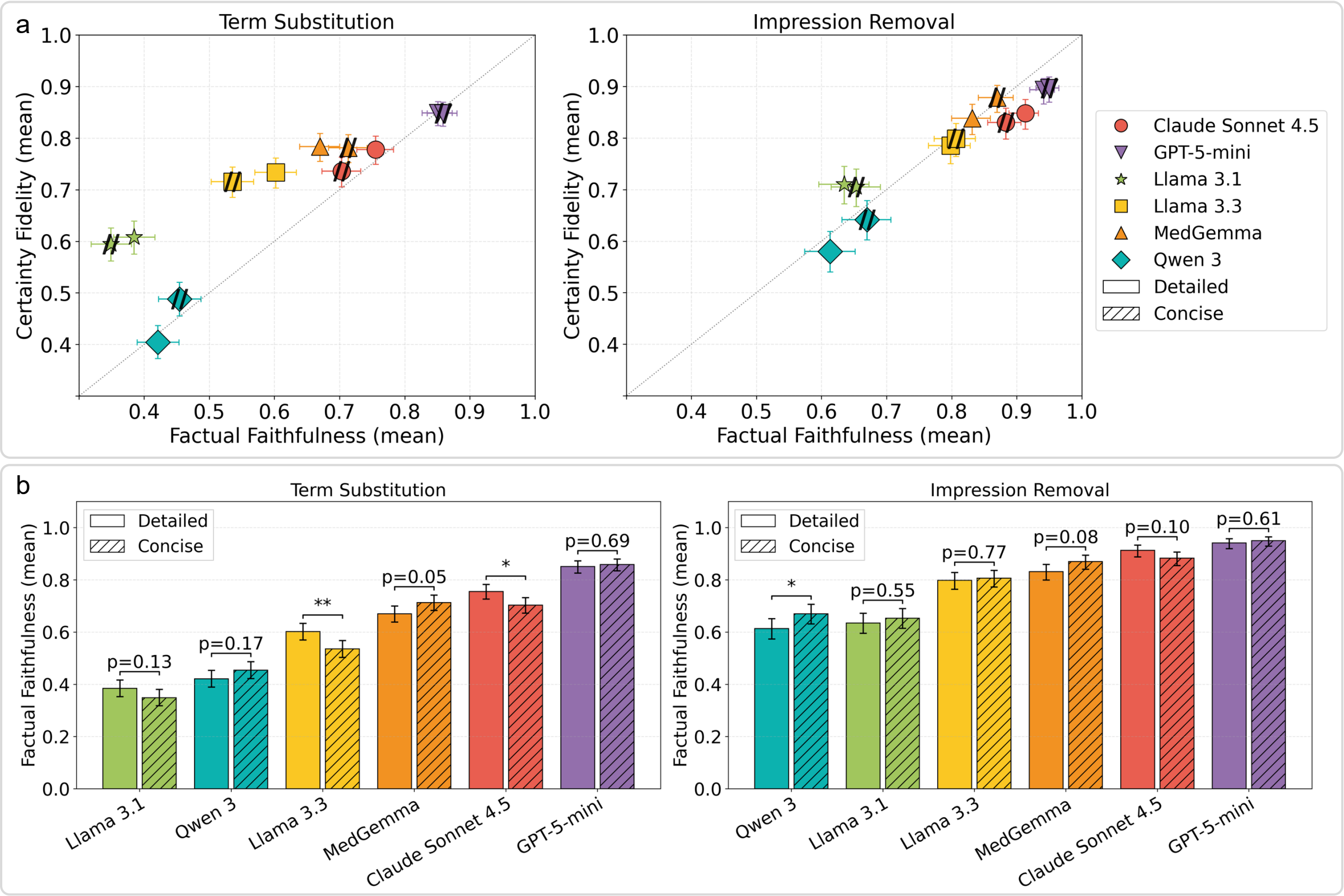

**Figure 3.** Comparison of summarizer performance on factual faithfulness and certainty fidelity across datasets and prompt styles. (a) Summarizers are compared based on their mean factual faithfulness (x-axis) and certainty fidelity (y-axis) under term-substitution and impression-removal adversarial edits. Detailed and Concise denote the two prompt variations used to generate the patient-friendly summary. Whiskers denote 95% Wilson score confidence intervals. GPT-5-mini achieves the highest factual faithfulness and certainty fidelity, followed by Claude Sonnet 4.5. (b) Bar plots show factual faithfulness stratified by detailed vs. concise prompting. Overall, GPT-5-mini achieves the highest factual faithfulness, while prompt differences have minor impact on the score across summarizers. Summary means were computed as proportions and 95% confidence intervals were calculated using the Wilson score interval. Nonparametric permutation tests within each summarizer and edit condition were used to compare Detailed vs. Concise prompts. (** denotes $p<0.01$, * denotes $p<0.05$).

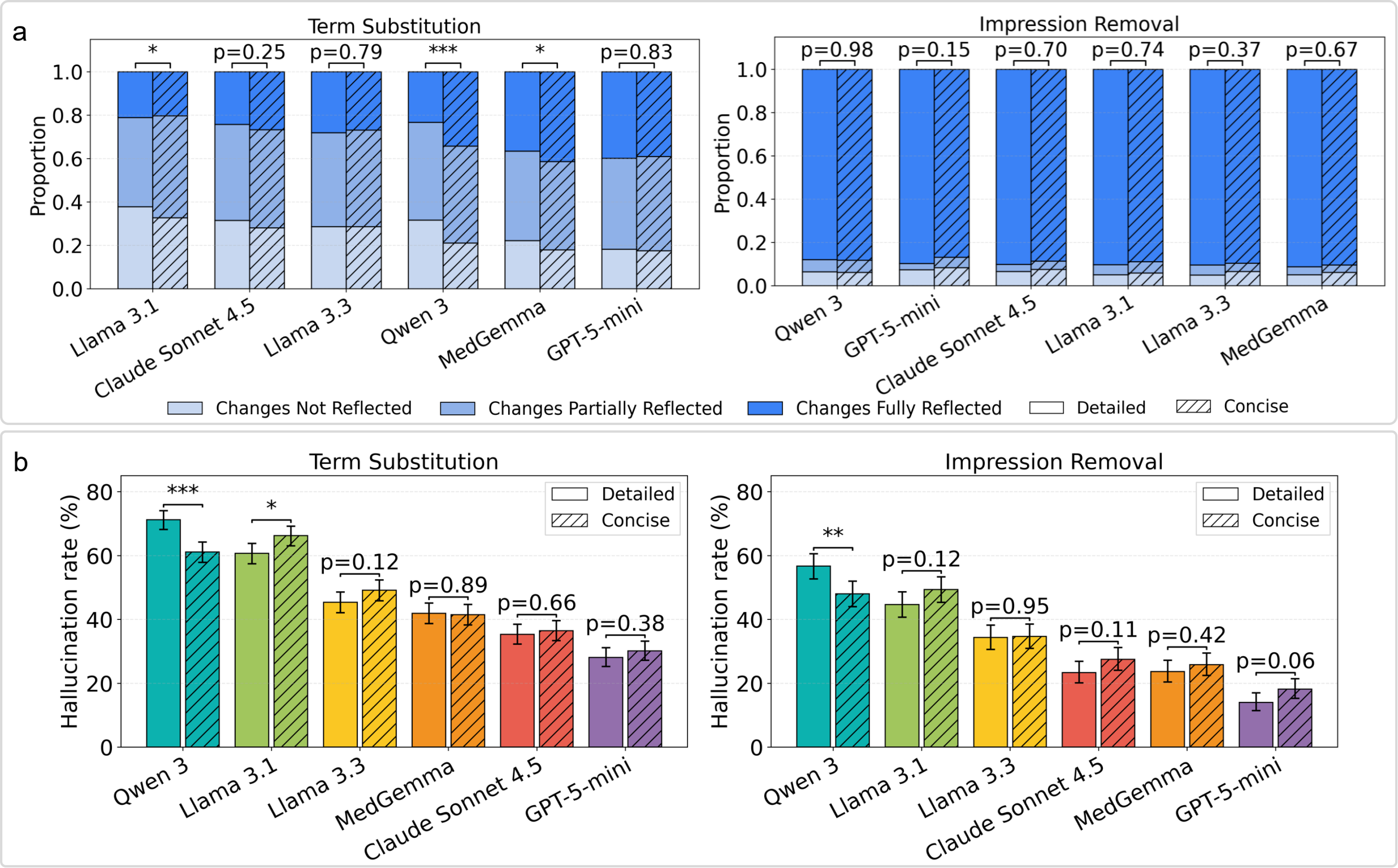

**Figure 4.** Sensitivity of summarizers to adversarial edits and hallucination rates across datasets. (a) Stacked bars display the proportion of summaries rated as fully sensitive, partially sensitive, or not sensitive to adversarial term substitutions and impression removal. Sensitivity is uniformly high for impression removal but more variable for term substitution. The patient-friendly summaries generated by GPT-5-mini and MedGemma reflect the term substitutions more reliably than the rest. (b) Hallucination rates (% of summaries containing findings, descriptions or recommendations absent from the radiology impressions) with Wilson 95% confidence intervals computed on the proportion of summaries containing hallucinations. Rates vary overall, with GPT-5-mini, MedGemma, and Claude Sonnet 4.5 showing the lowest rates across both adversarial conditions. Nonparametric permutation tests within each summarizer and edit condition were used to compare Detailed vs. Concise prompts. (*** denotes p<0.001, ** denotes p<0.01, * denotes p<0.05).

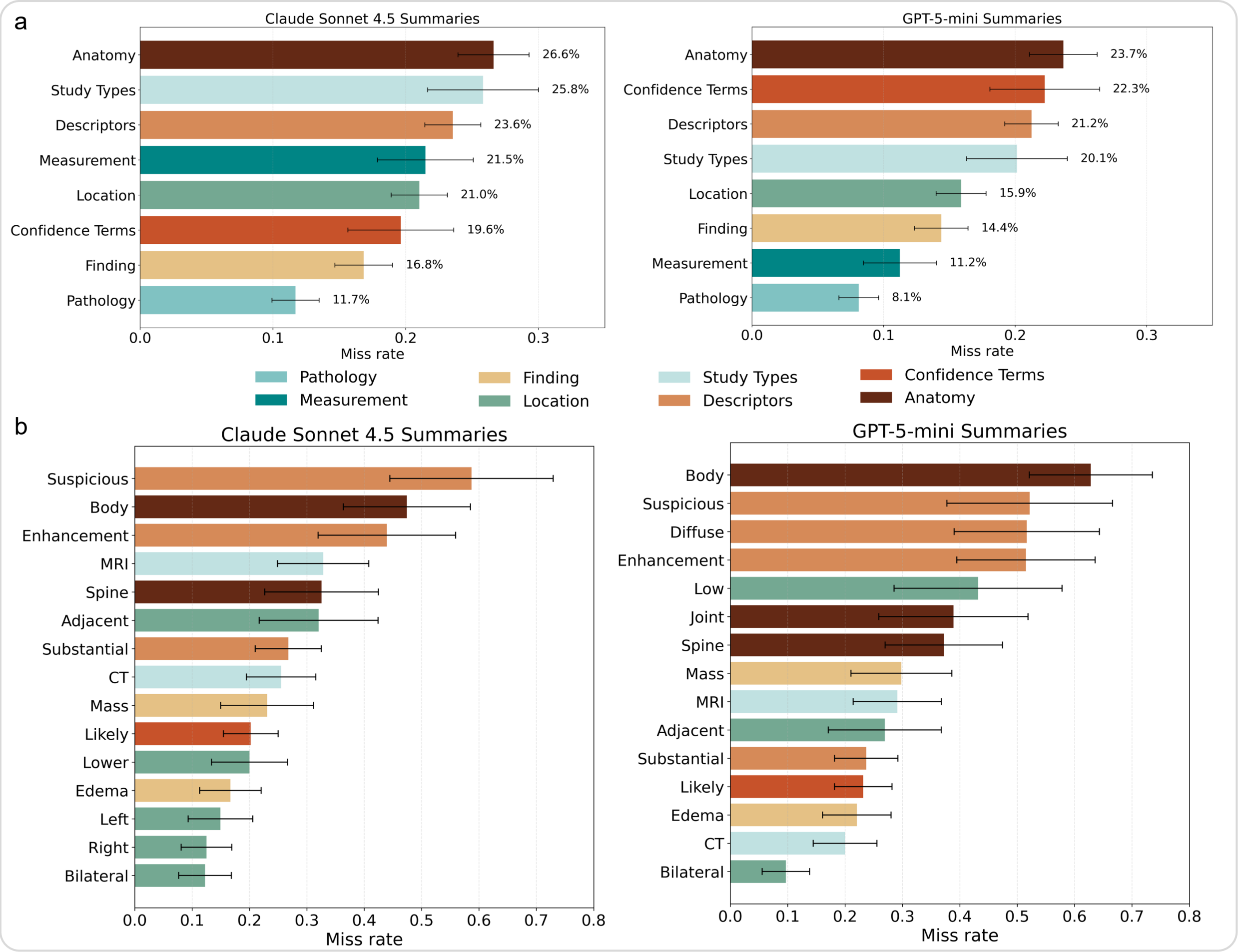

**Figure 5.** Failure mode analysis for adversarial term substitutions analyzed by category and individual terms. (a) Miss rates represent the percentage of adversarial substitutions within a specific category (e.g., Anatomy, Descriptors, Confidence Terms) that the summarizer failed to reflect in the patient-friendly summary, as detected by the GPT-5 Judge. Category-level miss rates show that terms from the categories of Anatomy, Confidence Terms and Descriptors are more frequently not reflected in the generated summaries, whereas pathology terms are least often missed. The pattern is consistent across the two highest-rated summarizers, Claude Sonnet 4.5 and GPT-5-mini. (b) Term-level analysis highlights specific substituted words with high miss rates, including descriptors such as “suspicious” and “substantial” and location-related terms, such as “adjacent” or “low”. These patterns identify systematic vulnerabilities in how summarizers incorporate clinically meaningful wording changes.

**Funding:**

R.G. acknowledges the German Academic Scholarship foundation.

R.D. acknowledges grant funding from ARPA-H LLM Care.

D.B.L. is a Member of the Board of Chancellors of the American College of Radiology and Board of Trustees of the American Board of Radiology, shareholder in Bunkerhill Health; and receives research funding from the Gordon and Betty Moore Foundation.

A.C. receives research support from NIH grants R01 HL167974, R01HL169345, R01 AR077604, R01 EB002524, R01 AR079431, P41 EB 027060, P50 HD118632; Advanced Research Projects Agency for Health (ARPA-H) Biomedical Data Fabric (BDF) and Chatbot Accuracy and Reliability Evaluation (CARE) programs (contracts AY2AX000045 and 1AYSAX0000024-01); and the Medical Imaging and Data Resource Center (MIDRC), which is funded by the National Institute of Biomedical Imaging and Bioengineering (NIBIB) under contract 75N92020C00021 and through ARPA-H.

Unrelated to this work, A.C. receives research support from GE Healthcare, Siemens, Philips, Microsoft, Amazon, Google, NVIDIA, Stability; has provided consulting services to Patient Square Capital, Chondrometrics GmbH, and Elucid Bioimaging; is co-founder of and receives personal fees from Cognita Imaging; has equity interest in Subtle Medical, LVIS Corp, Brain Key, and Radiology Partners.

**Acknowledgements:**

We thank Jason Wong for his feedback on the statistical analyses presented in this study.

# Supplementary Material

## 1. Construction of the Adversarial Term Dictionary

To create controlled and clinically meaningful perturbations for evaluating Large Language Model (LLM) robustness, we developed a systematic framework to construct an adversarial term dictionary. The methodology comprised three sequential stages: (1) automated medical term extraction, (2) similarity-based alternative generation, and (3) expert-guided refinement.

### Automated Medical Term Extraction

Medical terms were extracted from the "Impressions" section of over 13,000 reports from three distinct datasets: abdominal CTs (Merlin), chest CTPAs (INSPECT), and MRIs of the spine, brain, and other anatomies (MIMIC-IV). The extraction pipeline utilized a hybrid approach combining the scispaCy biomedical NLP library [1] for named entity recognition with rule-based patterns. After preprocessing steps to expand common abbreviations (e.g., "w/o" to "without"), the pipeline applied a series of regular expressions designed to capture specific radiological nomenclature. These patterns targeted complex pathologies (e.g., phrases ending in "syndrome," "disease," or "carcinoma"), staging and grading (Grade IV), measurements (5 mm), severity modifiers (mild, severe), and anatomical modifiers (bilateral, anterior). To ensure clinical relevance, candidate terms were filtered using a sentence-transformer model (pritamdeka/S-PubMedBert-MS-MARCO) [2]; only terms with a cosine similarity score greater than 0.4 to a curated set of medical domain concepts were retained. For example, given the impression “No pulmonary embolism. Mild bilateral pleural effusions. 1.5 cm right lower lobe nodule, suspicious for malignancy,” scispaCy identifies core biomedical entities such as “pulmonary embolism,” “pleural effusions,” and “right lower lobe nodule.” In parallel, rule-based patterns capture additional clinically meaningful modifiers and measurements, including “mild,” “bilateral,” “1.5 cm,” and “suspicious.” These entity- and pattern-derived candidates are consolidated and filtered via semantic similarity, yielding a final set of clinically relevant terms that jointly represent pathology, anatomy, severity, location, and uncertainty.

### Generation of Candidate Alternatives

For each extracted medical term, semantically plausible alternatives were generated using embedding-based similarity analysis. All terms were encoded using the S-PubMedBert-MS-MARCO model to create high-dimensional vector representations. A pairwise cosine similarity matrix was computed, and for each term, alternatives were selected from an empirically determined adaptive similarity band (typically the 50th-75th percentile). This approach identified terms that were semantically related but not direct synonyms. To ensure the quality of adversarial substitutions, a multi-step linguistic validation process was applied to filter out unsuitable candidates. This process excluded alternatives that were:

- **Semantically equivalent** (known synonyms or abbreviations).
- **Lexical overlaps** (substrings, morphological variants, or simple plurals).
- **Grammatically inconsistent** (did not share the same Part-of-Speech category as the original term).

**Dictionary Refinement and Validation**

The initial dictionary of alternatives underwent a final, automated refinement stage guided by a manually curated "helper" dictionary that defined term categories and relationships, guided by a radiologist. This refinement pipeline enforced several key rules:

- **Strict Category Enforcement:** Alternatives were pruned if they did not belong to the same semantic category (e.g., anatomy, pathology) as the base term.
- **Prioritization of Mutual Opposites:** For descriptors and confidence terms, the system prioritized alternatives with a bidirectional relationship in the helper dictionary (e.g., "mild"↔"severe"), ensuring a clear inversion of meaning.
- **Symmetry Enforcement:** For closed-set categories like study types, all members were made valid alternatives for one another.

The final dictionary was reviewed and validated by a $4^{th}$-year, board-eligible resident radiologist with expertise in computer vision and LLMs to ensure that substitutions could induce meaningful changes in a report's interpretation.

**Final Dictionary Characteristics**

The resulting adversarial dictionary contains 247 unique medical terms with 985 corresponding alternative mappings, averaging 3.99 alternatives per term. The terms are distributed across the following categories: pathology (57 terms), descriptors (63), anatomy (47), location (38), finding (25), confidence terms (9), and study types (8). The adversarial framework defines eight categories (main text, Figure 2); seven are represented in this dictionary, while the eighth, measurements, is altered numerically rather than by term substitution and therefore has no dictionary entries. This validated dictionary formed the basis for all subsequent adversarial edits used in our LLM evaluation framework. We release the adversarial dictionary used for all experiments in this study as Supplementary Data, a table listing all 247 terms and their 985 alternatives.

## 2. Prompts Used in the Study

### Prompts for Patient-Friendly Summary Generation

Two distinct prompts were used to generate patient-friendly summaries from the original and adversarially edited radiology impressions.

### Detailed Prompt

```
You are a medical communication specialist. Your task is to convert the following radiology impression into a clear, simple summary that a patient without medical background can understand.

Focus on:

   1. Explaining medical terms in plain language
   2. Highlighting what these findings mean for the patient
   3. Avoiding unnecessarily alarming language
   4. Being honest but reassuring where appropriate
   5. You will be evaluated in terms of completeness, conciseness,
      accuracy, and existence of hallucinations.

Radiology Impression:
{impression}

Patient-Friendly Explanation:
```

### Concise Prompt

```
Create a patient-friendly summary of the following radiology impression:
{impression}
```

## Prompts for LLM-as-a-Judge Evaluation

The automated evaluation framework utilized a two-stage prompting process to ensure a structured and objective assessment.

**Stage 1: Structured Fact Extraction Prompt**

```
You are a clinical NLP specialist. Your task is to extract structured
findings from the following radiology summary.

Each finding must be reported using the following fixed schema:

   – Finding ID: (start from 1, increment by 1)
   – Condition: (the medical condition being described, e.g.,
     "appendicitis")
   – Location: (the anatomical location, if any, e.g., "appendix",
     "lungs"; use "not specified" if missing)
   – Severity: (e.g., "mild", "moderate", "severe"; use "not applicable"
     if not stated)
   – Certainty: (must be exactly one of: "no evidence", "possible",
     "likely", "confirmed", "uncertain")
   – Differential: (if a differential diagnosis is mentioned, otherwise
     "not applicable")

Important Instructions:

   – If a field is not mentioned, fill it with the default (e.g., "not
     specified", "not applicable").
   – Certainty must be strictly one of the five listed values.
   – Do not infer differentials unless explicitly stated.
   – If there are no findings (e.g., "normal study"), respond: "No
     findings."
   – Use exactly the output layout below. No explanations or comments.

Output format:

Finding #1:

Condition:
Location:
Severity:
Certainty:
Differential (if any):

Finding #2: ...

Here is the radiology summary:
```

{title}:
{text}

**Stage 2: Comparative Judgment and Scoring Prompt**

```
You are a medical NLP evaluation expert. Your task is to compare extracted findings from four versions of a radiology summary:

   1. Original radiology report (Extracted Findings: {facts_A} )
   2. Adversarially modified radiology report (Extracted Findings: {facts_B})
   3. Patient-friendly summary of the original report (Simplified A: {simp_A})
   4. Patient-friendly summary of the adversarial report (Simplified B: {simp_B})

You are given:

   – Extracted structured findings from each version
   – Ground-truth alterations applied to create the adversarial report: {ground_truth_alterations}

Your task is to evaluate the *patient-friendly adversarial summary* by comparing it to the other three versions and the ground-truth alterations.

Score the summary on the following four categories:

---

### Category 1: Factual Faithfulness

Did the summary preserve all unaltered facts correctly?

→ Score: 0 (inaccurate), 1 (fully faithful)

### Category 2: Sensitivity to Changes

Did the summary reflect the adversarial term substitutions appropriately?

→ Score: 0 (missed), 1 (partially), or 2 (clearly reflected)

### Category 3: Certainty Fidelity

Did the summary correctly convey the diagnostic uncertainty (e.g., "no evidence" vs. "possible")?
```

```
→ Score: 0 (increased certainty), 1 (preserved or matched)

### Category 4: Hallucination Rate

Did the summary introduce any findings or details not present in any of
the 3 other versions?

→ Score: 0 (hallucinated), 1 (no hallucinations)

---

### Format your output strictly like this:

Factual Faithfulness: [0/1]
Sensitivity to Changes: [0/1/2]
Certainty Fidelity: [0/1]
Hallucination: [0/1]
Alterations: <e.g., ["acute -> healed", "obstruction -> stenosis"]>
Reasoning:

factual_faithfulness_reasoning: <explanation here>
sensitivity_to_changes_reasoning: <explanation here>
certainty_fidelity_reasoning: <explanation here>
hallucination_reasoning: <explanation here>

Important Notes:

– Use the extracted findings, not surface language alone.
– Focus on the *meaning* of alterations, not just word changes.
– Hallucination = introducing any new clinical findings not found in
  any of the 3 source documents.
– Certainty changes must be clinically meaningful (e.g., "possible
  pneumonia" vs. "pneumonia" matters).
```

## 3. Model Access Dates

**Table S1. Language Models Access Dates and API or serving information. For MIMIC-IV, we used Microsoft Azure to access GPT-5 and GPT-5-mini to comply with the PhysioNet MIMIC dataset requirements[1].**

| Model Name | Access Date | API / Serving |
| --- | --- | --- |
| **GPT-5 (gpt-5-2025-08-07) [3]** | 11/06/2025 | OpenAI API |
| **GPT 5-mini (gpt-5-mini-2025-08-07) [3]** | 10/02/2025 | OpenAI API |

[1] https://physionet.org/news/post/gpt-responsible-use/

| | | |
|---|---|---|
| **Claude Sonnet 4.5 [4]** | 10/02/2025 | Anthropic API |
| **Llama 3.1 [5]** | 06/10/2025 | vLLM [6], NVIDIA A6000 |
| **MedGemma [7]** | 09/29/2025 | vLLM [6], NVIDIA A6000 |
| **QWEN 3 [8]** | 09/09/2025 | Ollama[2], NVIDIA A6000 |
| **Llama 3.3 [5]** | 06/10/2025 | vLLM [6], NVIDIA A6000 |

## 4. MedGemma Judge Results

**Table S2. Inter-Rater Agreement Between MedGemma Judge, GPT-5 Judge, and Radiologists for Adversarial Patient-Friendly Summary Evaluation (n=105).** F1 scores represent symmetric macro F1 (mean ± SD from 800 bootstrap iterations).

| Metric | Rater Pair | F1-Score |
|---|---|---|
| **Factual Faithfulness** | MedGemma *vs.* Rad A | 0.46 ± 0.01 |
| | MedGemma *vs.* Rad B | 0.41 ± 0.03 |
| | MedGemma vs. GPT-5 | 0.40 ± 0.03 |
| **Certainty Fidelity** | MedGemma *vs.* Rad A | 0.57 ± 0.09 |
| | MedGemma *vs.* Rad B | 0.46 ± 0.01 |
| | MedGemma vs. GPT-5 | 0.46 ± 0.01 |
| **Sensitivity to Changes** | MedGemma *vs.* Rad A | 0.57 ± 0.05 |
| | MedGemma *vs.* Rad B | 0.49 ± 0.07 |
| | MedGemma vs. GPT-5 | 0.49 ± 0.06 |
| **Hallucination Rate** | MedGemma *vs.* Rad A | 0.43 ± 0.04 |
| | MedGemma *vs.* Rad B | 0.20 ± 0.04 |
| | MedGemma vs. GPT-5 | 0.34 ± 0.04 |
| **Overall** | MedGemma *vs.* Rad A | 0.51 ± 0.07 |
| | MedGemma *vs.* Rad B | 0.39 ± 0.13 |
| | MedGemma vs. GPT-5 | 0.42 ± 0.07 |

[2] https://ollama.com/

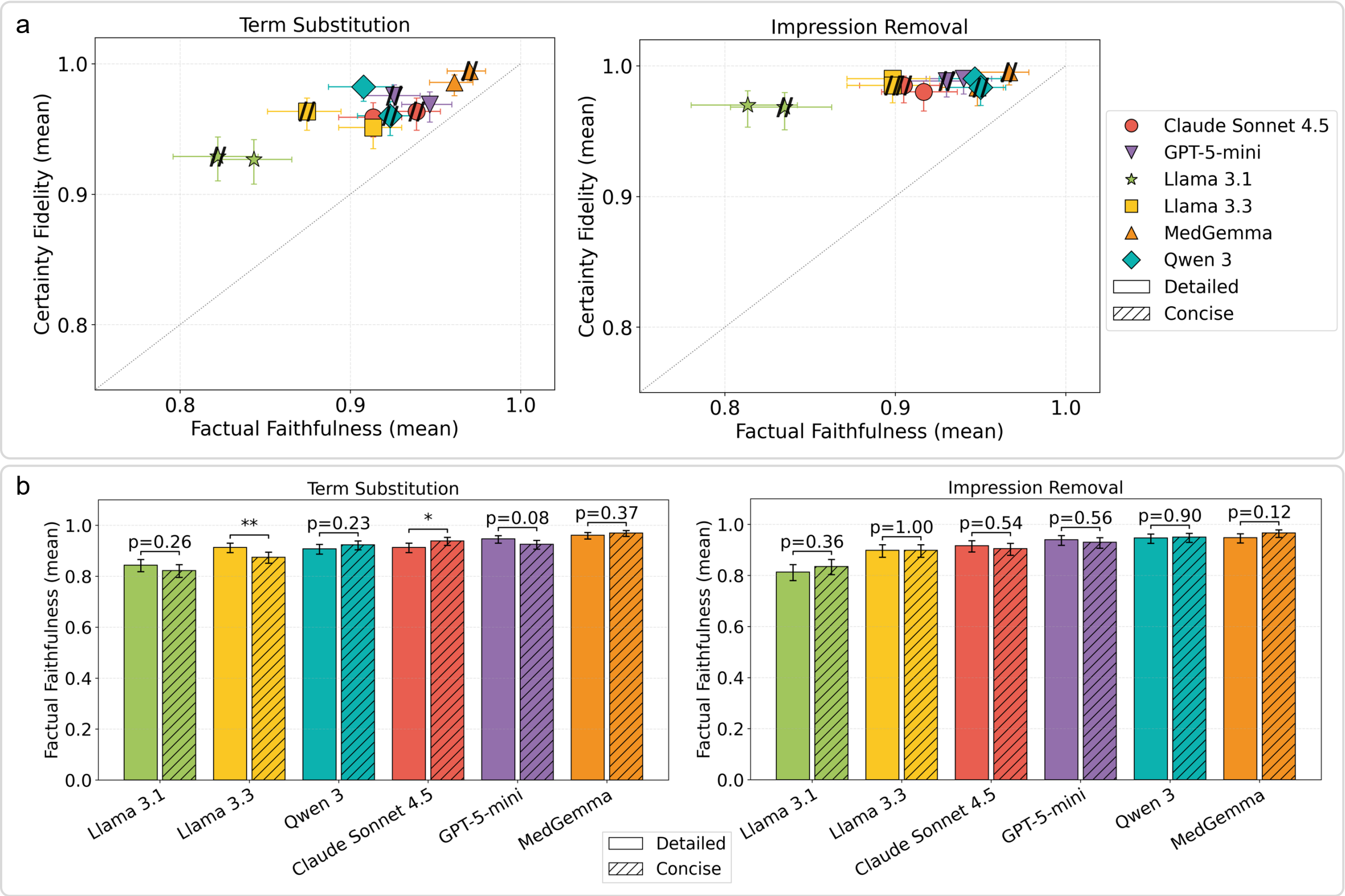

**Figure S1.** Comparison of summarizer performance on factual faithfulness and certainty fidelity across datasets and prompt styles based on scores assigned by MedGemma judge. (a) Summarizers are compared based on their mean factual faithfulness (x-axis) and certainty fidelity (y-axis) under term-substitution and impression-removal adversarial edits. Detailed and Concise denote the two prompt variations used to generate the patient-friendly summary. Whiskers denote 95% Wilson score confidence intervals. MedGemma achieves the highest factual faithfulness and certainty fidelity, followed closely by GPT-5-mini. (b) Bar plots show factual faithfulness stratified by detailed vs. concise prompting. Overall, MedGemma, GPT-5-mini and Qwen 3 achieve the highest factual faithfulness, while prompt differences have minor impact on the score across summarizers. Summary means were computed as proportions and 95% confidence intervals were calculated using the Wilson score interval. Nonparametric permutation tests within each summarizer and edit condition were used to compare Detailed vs. Concise prompts. (** denotes p<0.01, * denotes p<0.05).

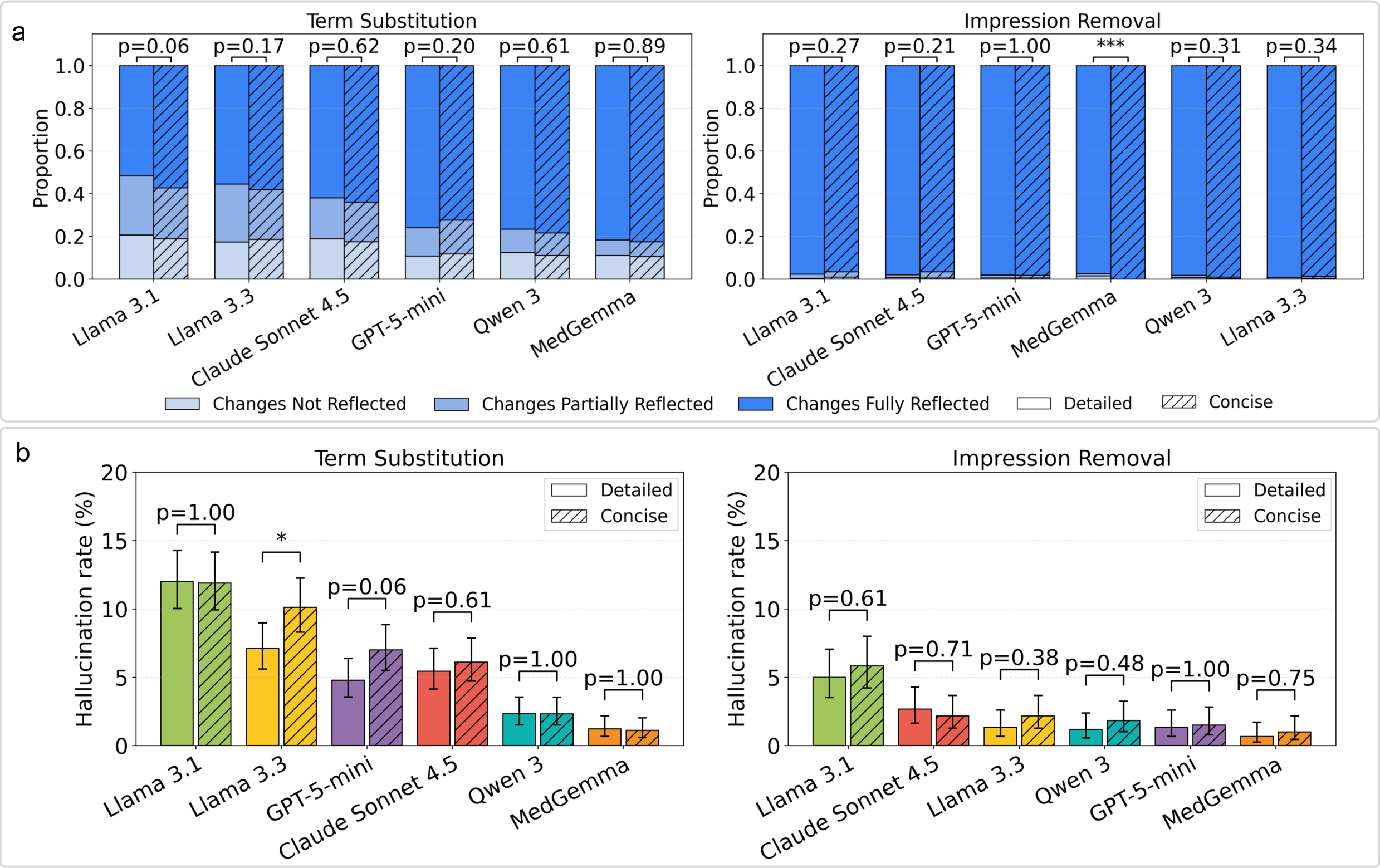

**Figure S2.** Sensitivity of summarizers to adversarial edits and hallucination rates across datasets based on scores assigned by MedGemma judge. (a) Stacked bars display the proportion of summaries rated as fully sensitive, partially sensitive, or not sensitive to adversarial term substitutions and impression removal. Sensitivity is uniformly high for impression removal but more variable for term substitution. The patient-friendly summaries generated by MedGemma, Qwen 3 and GPT-5-mini reflect the term substitutions more reliably than the rest. (b) Hallucination rates (% of summaries containing findings, descriptions or recommendations absent from the radiology impressions) with Wilson 95% confidence intervals computed on the proportion of summaries containing hallucinations. Rates vary overall, with MedGemma, GPT-5-mini, and Qwen 3 showing the lowest rates across both adversarial conditions. Nonparametric permutation tests within each summarizer and edit condition were used to compare Detailed vs. Concise prompts. (*** denotes $p<0.001$, ** denotes $p<0.01$, * denotes $p<0.05$).

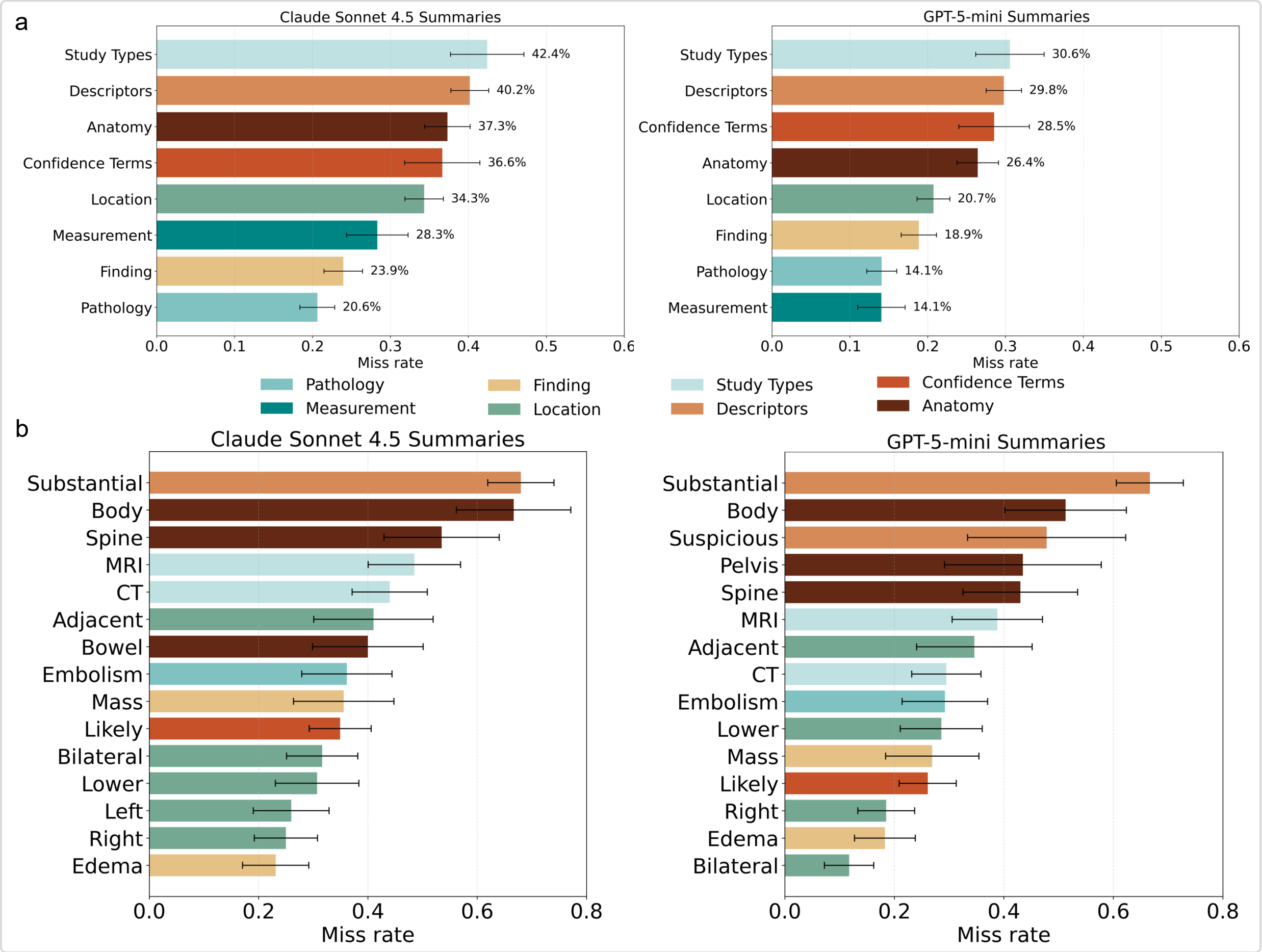

**Figure S3.** Failure mode analysis for adversarial term substitutions analyzed by category and individual terms based on MedGemma judge. (a) Miss rates represent the percentage of adversarial substitutions within a specific category (e.g., Anatomy, Descriptors, Confidence Terms) that the summarizer failed to reflect in the patient-friendly summary. Category-level miss rates show that terms from the categories of Study Types, Descriptors, Anatomy and Confidence Terms are more frequently not reflected in the generated summaries, whereas pathology terms and measurements are least often missed. The pattern is consistent across Claude Sonnet 4.5 and GPT-5-mini summarizers. (b) Term-level analysis highlights specific substituted words with high miss rates, including descriptors such as “substantial” and suspicious” and location-related terms, such as “adjacent” or “lower”. These patterns identify systematic vulnerabilities in how summarizers incorporate clinically meaningful wording changes.

## 5. Detailed Dataset Characteristics

**Table S3. Dataset characteristics.**

| Data-set | Mo-dality | Body region | Split | Filtering | Reports (N) | Impression points/report (mean±SD) | Words/impression section (mean±SD; median) |
|---|---|---|---|---|---|---|---|
| Merlin | CT | Abdomen | Official public patient-level test split | None (single-modality) | 5,125 | 2.7 ± 1.4 | 68.2 ± 49.2 (58) |
| INSPECT | CT (CTPA) | Chest | A priori patient-level test split (~16% of corpus) | None (single-modality) | 3,676 | 2.8 ± 1.5 | 63.9 ± 39.7 (57) |
| MIMIC-IV | MRI | Spine, brain/head, breast, knee, etc. | MIMIC-IV radiology corpus | MRI only; one report per patient; 5,000 longest impressions | 5,000 | 4.6 ± 2.2 | 149.2 ± 50.4 (143) |

## 6. Examples of summaries where scores differed across dimensions

We highlight two representative cases showing that the scoring dimensions are not redundant and that a summary can achieve high scores on some dimensions while failing on others. In Case 1, a summary scored factually faithful by the GPT-5 judge did not reflect the adversarial edit (sensitivity to changes failed); in Case 2, a summary that reflected the edit, preserved certainty, and introduced no new finding changed an unaltered fact (factual faithfulness failed). For each case, we summarize the edit, the relevant excerpts, and the four scores, reported as factual faithfulness / sensitivity to changes / certainty fidelity / hallucination rate. Sensitivity is scored 0, 1, or 2 (missed, partial, fully reflected); factual faithfulness and certainty fidelity are 0 or 1, and the hallucination rate is 1 when a hallucination is present and 0 when none is.

**Case 1.** Faithful summary that did not reflect the edit (INSPECT, Claude Sonnet 4.5 summarizer, concise prompt; scores 1 / 0 / 1 / 0). The adversarial edit escalated the right hilar lymphadenopathy from “mild” to “marked”. The summary preserved the finding faithfully but reported only “enlarged lymph nodes near your right lung”, without conveying the increase in severity, so the escalation to “marked” did not reach the patient. Because the altered term was not reflected, sensitivity to changes was scored as 0, while factual faithfulness was scored as 1.

**Case 2.** Summary that reflected the edit but also changed an unaltered fact (INSPECT, Llama 3.3 summarizer, detailed prompt; scores 0 / 2 / 1 / 0). The adversarial edit changed the pulmonary emboli from bilateral to unilateral, and the summary tracked this exactly (“both lungs” became “one of your lungs”) while keeping the report’s hedging about possibly raised right-heart pressure. But it misstated an unaltered fact, placing the clots in “the smaller airways” rather than in the blood vessels (the pulmonary arteries). Because an unaltered fact was not preserved, factual faithfulness was scored 0, even though the summary reflected the edit (sensitivity 2), preserved certainty, and introduced no new finding.

## 7. Supplementary Statistical Analysis (GPT-5 Judge)

**Methods.** To exploit the factorial structure of the evaluation, we fitted a single multi-factor model per metric. The design is a balanced repeated-measures factorial in which each of 300 reports (100 per dataset) was evaluated under all combinations of summarizer (6), prompt style (2), adversarial condition (5: term substitution at easy/medium/hard intensity and impression removal at easy/hard intensity), and dataset (3), giving 18,000 evaluations per metric and 60 per report. Because edit type and intensity are not fully crossed, they were combined into a single five-level condition factor. The three binary metrics (factual faithfulness, certainty fidelity, hallucination) were analysed with population-averaged logistic regression fitted by generalised estimating equations, using an exchangeable working correlation with reports as clusters and robust sandwich standard errors. Sensitivity to changes, an ordinal 0/1/2 score, was analysed with a proportional-odds cumulative logit model with report-clustered sandwich standard errors. Each model included all four factors and all two-way interactions; factors were sum-to-zero coded so that Wald tests of lower-order terms are Type III tests. An omnibus Type III Wald test was reported for every term before any pairwise comparison, and pairwise contrasts were estimated from the fitted model as linear combinations on the link scale with Holm adjustment within each metric. Effects are reported as odds ratios with 95% confidence intervals, averaged with equal weight over the cells of the balanced design, so they do not depend on the coding of the factors. Marginal estimates are reported both as link-scale estimated marginal means and as marginally standardised average fitted responses, with observed differences computed per report as a model-free check. Because the proportional-odds assumption was imperfect, sensitivity was also modelled as separate P(score≥1) and P(score≥2) logistic models, and threshold odds ratios were compared using the joint report-clustered covariance of the two models. A three-way summarizer × prompt × condition term was fitted as a sensitivity analysis, and random-intercept logistic mixed models (random intercept per report) as a consistency check. Analyses used Python 3.10 with statsmodels 0.14.6; two-sided $\alpha = 0.05$.

**Results.** The two-way interactions were jointly significant for all four metrics (49 df, all $p < 10^{-19}$), so main effects alone do not describe the data adequately. Summarizer, adversarial condition, and dataset had strong main effects (all $p < 10^{-15}$). Averaged over all design cells, the detailed versus concise prompt did not significantly change the odds of any outcome (factual faithfulness odds ratio 0.99, 95% CI 0.92-1.06; certainty fidelity 0.97, 0.91-1.04; hallucination 0.97, 0.90-1.04; sensitivity 0.96, 0.91-1.02). The average probability of the binary outcomes also did not change significantly, but the mean sensitivity score was slightly lower with the detailed prompt (−0.031 on the 0-2 scale, 95% CI −0.043 to −0.018), because the effect reversed by edit type (term substitution odds ratio 0.84, 95% CI 0.80-0.88; impression removal 1.17, 1.04-1.32). Prompt style interacted significantly with summarizer for all four metrics ($p = 1.2\times10^{-10}$ to $1.4\times10^{-5}$), indicating that its effect is model-specific. Qwen 3 was the only summarizer for which the detailed prompt was significantly worse on all four metrics (factual faithfulness odds ratio 0.81, 95% CI 0.70-0.94; certainty fidelity 0.71, 0.62-0.80; hallucination 1.52, 1.31-1.77; sensitivity 0.69, 0.62-0.78). The proportional-odds assumption was imperfect for sensitivity: harder term substitution raised the odds of an edit being at least partly reflected (hard vs easy odds ratio 13.63, 95% CI 10.41-17.83) but lowered the odds of it being fully reflected (0.55, 0.46-0.67). Report-level intraclass correlations of 0.086-0.213 confirmed substantial clustering of judge scores within reports. Model-based marginal estimates agreed with the descriptive rates in the main text to within 0.002 for the binary metrics.

**Table S4. Type III Wald tests (M2: main effects + all two-way interactions), p-values.**

| Term | df | Factual faithfulness | Certainty fidelity | Hallucination | Sensitivity |
|---|---|---|---|---|---|
| Summarizer | 5 | $1.1\times10^{-275}$ | $2.9\times10^{-169}$ | $5.1\times10^{-222}$ | $5.9\times10^{-22}$ |
| Prompt | 1 | 0.733 | 0.426 | 0.334 | 0.150 |
| Condition | 4 | $6.6\times10^{-56}$ | $2.8\times10^{-28}$ | $7.6\times10^{-58}$ | $9.8\times10^{-90}$ |
| Dataset | 2 | $2.1\times10^{-65}$ | $2.6\times10^{-16}$ | $1.1\times10^{-26}$ | $5.1\times10^{-21}$ |
| Summarizer × Prompt | 5 | $6.5\times10^{-7}$ | $1.4\times10^{-5}$ | $2.2\times10^{-8}$ | $1.2\times10^{-10}$ |
| Summarizer × Condition | 20 | 0.171 | $3.7\times10^{-6}$ | 0.003 | $2.4\times10^{-13}$ |
| Summarizer × Dataset | 10 | $1.4\times10^{-5}$ | $6.8\times10^{-9}$ | $3.8\times10^{-4}$ | 0.113 |
| Prompt × Condition | 4 | 0.079 | 0.250 | 0.592 | $1.7\times10^{-5}$ |
| Prompt × Dataset | 2 | $8.9\times10^{-4}$ | 0.825 | 0.004 | 0.097 |
| Condition × Dataset | 8 | 0.003 | $1.8\times10^{-4}$ | $1.3\times10^{-13}$ | 0.002 |

**Table S5. Odds ratios (95% CI) for the prompt effect and the main effects, averaged with equal weight over the other factors.**

| Contrast | Factual faithfulness | Certainty fidelity | Hallucination | Sensitivity |
|---|---|---|---|---|
| **Prompt, detailed vs concise** | | | | |
| Average over all cells | 0.99 (0.92-1.06) | 0.97 (0.91-1.04) | 0.97 (0.90-1.04) | 0.96 (0.91-1.02) |
| Within GPT-5-mini | 0.89 (0.69-1.13) | 1.00 (0.82-1.21) | 0.85 (0.71-1.03) | 1.14 (1.02-1.28) |
| Within Claude Sonnet 4.5 | **1.33 (1.11-1.60)** | 1.24 (1.05-1.46) | 0.90 (0.76-1.06) | 0.98 (0.88-1.08) |
| Within MedGemma 27B | **0.76 (0.63-0.91)** | 0.90 (0.76-1.07) | 0.98 (0.83-1.14) | 0.90 (0.80-1.01) |

| Contrast | Factual faithfulness | Certainty fidelity | Hallucination | Sensitivity |
|---|---|---|---|---|
| Within Llama 3.3 (70B) | 1.20 (1.01-1.41) | 1.03 (0.88-1.20) | 0.90 (0.77-1.05) | 1.13 (1.00-1.27) |
| Within Qwen 3 (4B) | **0.81 (0.70-0.94)** | **0.71 (0.62-0.80)** | **1.52 (1.31-1.77)** | **0.69 (0.62-0.78)** |
| Within Llama 3.1 (8B) | 1.07 (0.93-1.24) | 1.05 (0.90-1.21) | **0.80 (0.68-0.93)** | 0.99 (0.88-1.12) |
| Within term substitution | 1.06 (0.97-1.16) | 1.01 (0.93-1.10) | 0.99 (0.91-1.07) | **0.84 (0.80-0.88)** |
| Within impression removal | 0.88 (0.79-0.99) | 0.92 (0.83-1.02) | 0.94 (0.84-1.04) | **1.17 (1.04-1.32)** |
| Within INSPECT | 1.13 (0.99-1.29) | 1.00 (0.89-1.13) | 0.98 (0.87-1.10) | 1.02 (0.94-1.11) |
| Within Merlin | **0.83 (0.74-0.93)** | 0.96 (0.85-1.07) | 1.10 (0.99-1.23) | 0.95 (0.87-1.03) |
| Within MIMIC-IV | 1.02 (0.90-1.17) | 0.96 (0.87-1.07) | **0.84 (0.75-0.94)** | 0.91 (0.84-0.99) |
| **Summarizer vs GPT-5-mini** | | | | |
| Claude Sonnet 4.5 | **0.48 (0.39-0.58)** | **0.60 (0.52-0.70)** | **1.53 (1.32-1.77)** | **0.70 (0.63-0.79)** |
| MedGemma 27B | **0.35 (0.29-0.42)** | **0.66 (0.56-0.77)** | **1.70 (1.50-1.94)** | 1.09 (0.97-1.22) |
| Llama 3.3 (70B) | **0.21 (0.17-0.25)** | **0.46 (0.39-0.53)** | **2.50 (2.18-2.87)** | **0.76 (0.67-0.85)** |
| Qwen 3 (4B) | **0.10 (0.08-0.12)** | **0.15 (0.13-0.18)** | **5.56 (4.88-6.33)** | **0.74 (0.65-0.84)** |
| Llama 3.1 (8B) | **0.08 (0.07-0.10)** | **0.27 (0.23-0.32)** | **4.70 (4.08-5.42)** | **0.61 (0.53-0.70)** |
| **Edit type and intensity** | | | | |
| Term substitution vs impression removal | **0.32 (0.28-0.37)** | **0.58 (0.51-0.67)** | **2.01 (1.82-2.23)** | **0.05 (0.03-0.06)** |
| Term substitution: medium vs easy ‡ | 0.92 (0.82-1.04) | **0.82 (0.73-0.92)** | **1.25 (1.13-1.39)** | **1.28 (1.03-1.59)** |
| Term substitution: hard vs medium ‡ | **0.85 (0.75-0.96)** | **0.74 (0.66-0.84)** | **1.48 (1.32-1.66)** | **1.37 (1.22-1.55)** |
| Term substitution: hard vs easy ‡ | **0.78 (0.69-0.88)** | **0.61 (0.53-0.70)** | **1.86 (1.66-2.08)** | **1.76 (1.45-2.14)** |
| Impression removal: hard vs easy | **1.28 (1.09-1.49)** | **1.27 (1.09-1.48)** | 0.93 (0.83-1.03) | 1.28 (0.96-1.69) |
| **Dataset** | | | | |
| INSPECT vs Merlin | 1.23 (0.96-1.58) | 0.99 (0.75-1.30) | **0.69 (0.58-0.83)** | 1.34 (0.96-1.87) |
| INSPECT vs MIMIC-IV | **4.64 (3.82-5.64)** | **2.37 (1.86-3.03)** | **0.42 (0.36-0.50)** | **3.44 (2.61-4.55)** |
| Merlin vs MIMIC-IV | **3.77 (3.03-4.69)** | **2.40 (1.88-3.05)** | **0.61 (0.51-0.72)** | **2.57 (1.94-3.41)** |

OR > 1 means a higher score, which is favourable except for hallucination (1 = present). Sensitivity is from the proportional-odds model. p-values for each term are in Table S4. Bold: Holm-adjusted p<0.05 within the family of contrasts (each prompt grouping; all 15 summarizer pairs; the 5 edit-type and intensity contrasts; the 3 dataset pairs). Grey text: the parent Type III term is not significant, so the value is not interpreted. ‡: in the Sensitivity column the pooled OR is not interpretable; see Table S6.

**Table S6. Proportional-odds diagnostic for the ordinal sensitivity model: odds ratios (95% CI) under the pooled model and at each threshold (slope correlation across splits r = 0.54).**

| Contrast | Pooled OR | P(Y≥1) OR | P(Y≥2) OR | Ratio, P(Y≥1) / P(Y≥2) | Ratio p |
|---|---|---|---|---|---|
| Prompt: average over all cells | 0.96 (0.91-1.02) | 0.91 (0.84-0.99) | 0.98 (0.92-1.04) | 0.93 (0.86-1.01) | 0.075 |
| Prompt: within term substitution | **0.84 (0.80-0.88)** | **0.75 (0.68-0.82)** | **0.88 (0.82-0.94)** | 0.85 (0.76-0.94) | 0.002 |
| Prompt: within impression removal | **1.17 (1.04-1.32)** | **1.22 (1.04-1.42)** | **1.14 (1.02-1.29)** | 1.06 (0.93-1.22) | 0.369 |
| Term substitution vs impression removal | **0.05 (0.03-0.06)** | **0.23 (0.17-0.32)** | **0.04 (0.03-0.05)** | 5.97 (5.06-7.05) | $2.6\times10^{-99}$ |
| Term substitution: medium vs easy | **1.28 (1.03-1.59)** | **2.63 (2.11-3.26)** | **0.57 (0.47-0.70)** | 4.59 (3.80-5.55) | $4.4\times10^{-56}$ |
| Term substitution: hard vs medium | **1.37 (1.22-1.55)** | **5.19 (3.89-6.92)** | 0.97 (0.81-1.16) | 5.35 (4.00-7.18) | $2.9\times10^{-29}$ |
| Term substitution: hard vs easy | **1.76 (1.45-2.14)** | **13.63 (10.41-17.83)** | **0.55 (0.46-0.67)** | 24.60 (19.01-31.84) | $8.0\times10^{-131}$ |
| Impression removal: hard vs easy | 1.28 (0.96-1.69) | **1.42 (1.04-1.94)** | 1.24 (0.95-1.63) | 1.14 (0.97-1.34) | 0.103 |

P(Y≥1): at least partly reflected, vs not reflected. P(Y≥2): fully reflected, vs at most partly reflected. Bold: Holm-adjusted p<0.05, across the model terms for the average prompt row and within each family of contrasts otherwise. Ratio: the P(Y≥1) OR divided by the P(Y≥2) OR, from the joint report-clustered covariance of the two models, unadjusted; 1 means the same effect at both thresholds.